\pdfoutput=1

\documentclass[11pt]{article}

\usepackage{acl}
\usepackage[utf8]{inputenc}
\usepackage{arydshln}
\usepackage{dirtytalk}
\usepackage{svg}

\usepackage{soul}
\usepackage{xcolor}
\usepackage{times}
\usepackage{latexsym}
\usepackage{comment}
\usepackage{multirow}
\usepackage{booktabs}
\usepackage{subcaption}
\usepackage{tipa}
\usepackage{bbm}
\usepackage{amsmath,amssymb}
\DeclareMathOperator*{\argmax}{arg\,max}

\usepackage{tikz}
\usetikzlibrary{positioning, fit, shapes}

\DeclareMathOperator{\E}{\mathbb{E}}
\usepackage{graphicx}
\usepackage{mathtools}

\usepackage{xcolor}
\usepackage{hyperref}
\usepackage{bm}

\newcommand{\note}[2]{{\color{#1}{#2}}}
\newcommand{\todo}[1]{\note{red}{\textbf{TODO:}[#1]}}

\providecommand{\key}[1]{\textbf{#1}}

\renewcommand{\todo}[1]{\ignorespaces}

\usepackage[T1]{fontenc}

\usepackage[utf8]{inputenc}

\usepackage{microtype}
\usepackage{xspace}
\usepackage{mathtools}
\usepackage[capitalize]{cleveref}

\usepackage{algorithm2e}
\RestyleAlgo{ruled}
\SetKwInput{KwParameters}{Parameters}

\crefname{algocf}{alg.}{algs.}
\Crefname{algocf}{Algorithm}{Algorithms}

\newcommand{\ie}{\emph{i.e.,}\xspace}

\newcommand{\entropymath}{\mathcal{H}}
\newcommand{\entropy}{$\mathcal{H}$\xspace}

\newcommand{\candpooltrainmath}{\Language}
\newcommand{\candpooltrain}{\Language}

\newcommand{\candpoolrandom}{$\mathcal{X}$\xspace}

\newcommand{\candpoolrandommath}{\mathcal{X}}

\newcommand{\policy}{\ensuremath{\pi}}
\newcommand{\atrain}{$\pi_\text{train}$\xspace}
\newcommand{\aunif}{$\pi_\text{unif}$\xspace}

\newcommand{\ainfo}{$\pi_\text{eig}$\xspace}

\newcommand{\alabelent}{$\pi_\text{label-ent}$\xspace}

\newcommand{\Xmath}{\underline{\smash{x}}}
\newcommand{\Ymath}{\underline{\smash{y}}}
\newcommand{\policyinputmath}{x|\Xmath,\Ymath}

\newcommand{\atrainmath}{\pi_\text{train}}

\newcommand{\aunifmath}{\pi_\text{unif}}

\newcommand{\ainfomath}{\pi_\text{eig}}

\newcommand{\alabelentmath}{\pi_\text{label-ent}}

\newcommand{\SelectedString}{\ensuremath{x^*}}

\newcommand{\infotrainhistory}{$\pi_\text{eig-history}$\xspace}
\newcommand{\infotrainmodel}{$\pi_\text{eig-model}$\xspace}
\newcommand{\infotrainmixed}{$\pi_\text{eig-mixed}$\xspace}

\newcommand{\infogaintrainhistorymath}{\pi_\text{eig-history}}

\newcommand{\metricinfogain}{$V_{\text{IG}}$\xspace}

\newcommand{\metricinfogainmath}{V_{\text{IG}}}
\newcommand{\metricinfogainmathbold}{\bm{V}_{\textbf{IG}}}

\newcommand{\scoreempiricalmath}{S^{\text{EMP}}}
\newcommand{\scoreempiricalmathbold}{\bm{S}^{\textbf{EMP}}}

\newcommand{\scoremodeltrainmath}{S^{\text{EXP}{(x)}}}
\newcommand{\scoremodeltrainmathbold}{\bm{S}^{\textbf{EXP}\bm{(x)}}}

\newcommand{\scoremodelnontrainmath}{S^{\text{EXP}{(y)}}}
\newcommand{\scoremodelnontrainmathbold}{\bm{S}^{\textbf{EXP}\bm{(y)}}}

\newcommand{\nontrain}{$\hat{\pi}$\xspace}
\newcommand{\nontrainmath}{\hat{\pi}}

\newcommand{\fulltheta}{$\boldsymbol{\theta}$\xspace}
\newcommand{\fullthetamath}{\boldsymbol{\theta}}

\newcommand{\priormath}{\theta_{\text{prior}}}

\newcommand\sect[1]{\S\ref{#1}}

\usepackage{float}

\title{Learning Phonotactics from Linguistic Informants}

\makeatletter
\renewcommand*{\@fnsymbol}[1]{\ifcase#1\or\dagger\else\@arabic{#1}\fi}
\makeatother

\author{
{Canaan Breiss\textsuperscript{a,}\thanks{\;\,Both authors contributed equally to this work.}\quad Alexis Ross\textsuperscript{b,}\footnotemark[1]}
\\
{\textbf{Amani Maina-Kilaas\textsuperscript{b} \quad Roger Levy\textsuperscript{b} \quad Jacob Andreas\textsuperscript{b}}}\\
 {\textsuperscript{a}University of Southern California \quad \textsuperscript{b}Massachusetts Institute of Technology}\\
\texttt{cbreiss@usc.edu} \quad \texttt{\{alexisro, amanirmk, rplevy, jda\}@mit.edu}\\
}

\begin{document}
\maketitle
\begin{abstract}
We propose an interactive approach to language learning that utilizes linguistic acceptability judgments from an informant (a competent language user) to learn a grammar. Given a grammar formalism and a framework for synthesizing data, our model iteratively selects or synthesizes a data-point according to one of a range of information-theoretic policies, asks the informant for a binary judgment, and updates its own parameters in preparation for the next query. We demonstrate the effectiveness of our model in the domain of phonotactics, the rules governing what kinds of sound-sequences are acceptable in a language, and carry out two experiments, one with typologically-natural linguistic data and another with a range of procedurally-generated languages. We find that the information-theoretic policies that our model uses to select items to query the informant achieve sample efficiency comparable to, and sometimes greater than, fully supervised approaches.

\end{abstract}

\section{Introduction}

In recent years, natural language processing has made remarkable progress toward models that can (explicitly or implicitly) predict and use representations of linguistic structure from phonetics to syntax \citep{mohamed2022self,hewitt2019structural}.
These models play a prominent role in contemporary computational linguistics research. But the data required to train them is of a vastly larger scale, and features less controlled coverage of important phenomena, than data gathered in the course of linguistic research, e.g.\ during language documentation with native speaker informants. How can we build computational models that learn more \emph{like linguists}---from targeted inquiry rather than large-scale corpus data?

We describe a paradigm in which language-learning agents interactively select examples to learn from by querying an \textbf{informant}, with the goal of learning about a language as data-efficiently as possible, rather than relying on large-scale corpora to capture attested-but-rare phenomena. 
This approach has two important features. First, rather than relying on existing data to learn, our model performs \textbf{data synthesis} to explore the space of useful possible data-points. But second, our model can also \textbf{leverage corpus data} as part of its learning procedure by trading off between interactive elicitation and ordinary supervised learning, making it useful both \textit{ab initio} and in scenarios where seed data is available to bootstrap a full grammar. %

\begin{figure}
    \includegraphics[width=\columnwidth]{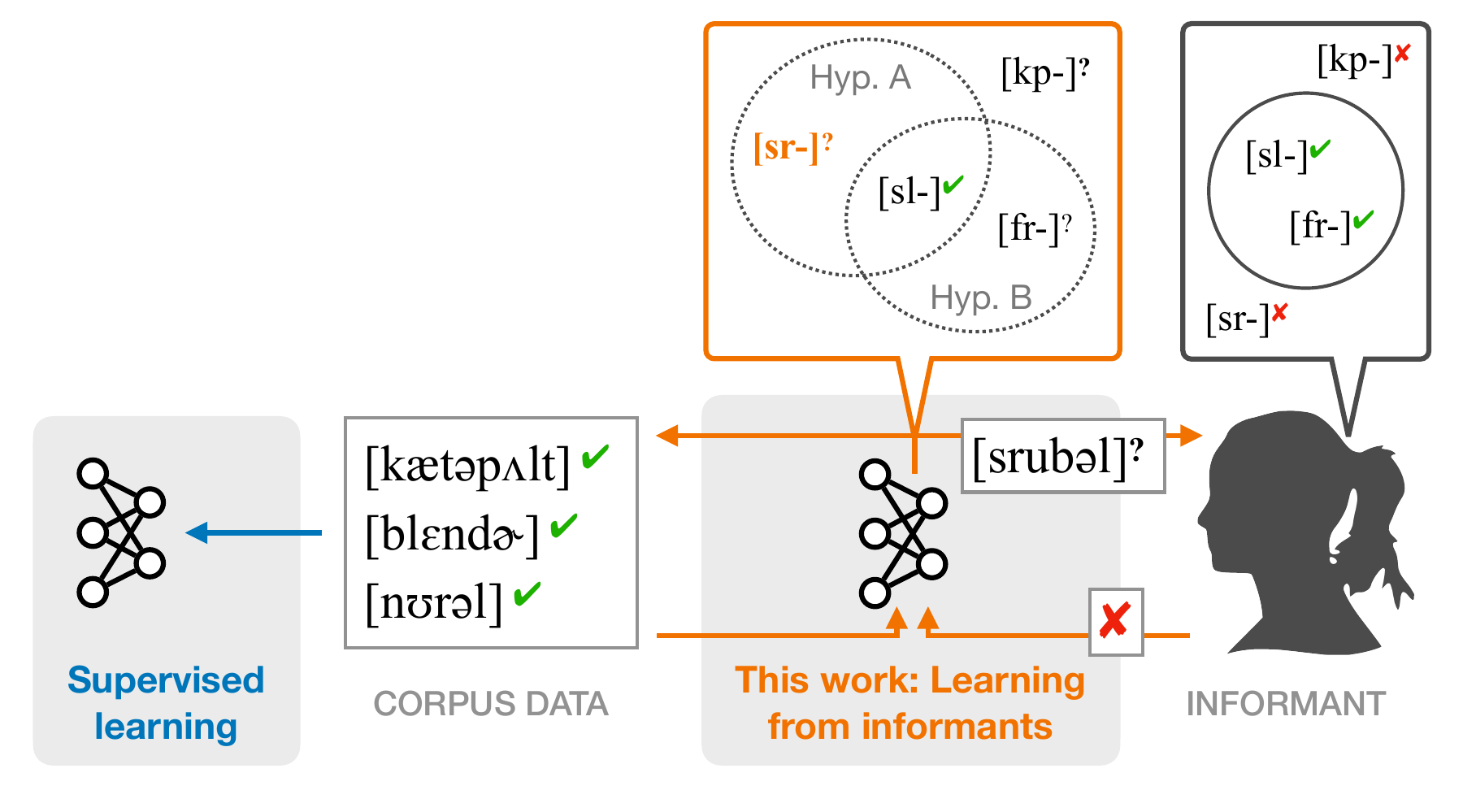} %
    \caption{Overview of our approach. Instead of learning a model from a static set of well-formed word forms (left), we interactively elicit acceptability judgments from a knowledgeable language user (right), using ideas from active learning and optimal experiment design. On a family of phonotactic grammar learning problems, active example selection is sometimes more sample-efficient than supervised learning or elicitation of judgments about random word forms.}
    \label{fig:teaser}
\end{figure}

We evaluate the capabilities of our methods in two experiments on learning \textit{phonotactic grammars}, in which the goal is to learn the constraints on sequences of permissible sounds in the words of a language. Applied to the problem of learning a vowel harmony system inspired by natural language typology, we show that our approach succeeds in recovering the generalizations governing the distribution of vowels. Using an additional set of procedurally-generated synthetic languages, our approach also succeeds in recovering relevant phonotactic generalizations, demonstrating that model performance is robust to whether the target pattern is typologically common or not. 
We find that our approach is more sample-efficient than ordinary supervised learning or random queries to the informant.

Our methods have the potential to be deployed as an aid to learners acquiring a second language or to linguists doing elicitation work with speakers of a language that has not previously been documented. Further, the development of more data-efficient computational models can help redress social inequalities which flow from the asymmetrical distribution of training data types available for present models \citep{bender2021dangers}.

\section{Problem Formulation and Method}
\label{sec:problem}

\providecommand{\Alphabet}{\ensuremath{\Sigma}}
\providecommand{\SigmaStar}{\ensuremath{\Alphabet^+}}
\providecommand{\Language}{\ensuremath{L}}

\paragraph{Preliminaries}
We aim to learn a language $\Language$ comprising a set of strings $x$, each of which is a concatenation of symbols from some inventory $\Alphabet$ (so $\Language \subseteq \SigmaStar$). 
(In phonotactics, for example, $\Alphabet$ might be the set of phonemes, and $\Language$ the set of word forms that speakers judge phonotactically acceptable.)
A learned model of a language is a 
discriminative function that maps from elements $x \in \SigmaStar$ to values in $\{0,1\}$ where 1 indicates that $x \in \Language$ and 0 indicates that $x \notin \Language$. In this paper, we will generalize this to \textbf{graded} models of language membership $f: \SigmaStar \mapsto [0,1]$, in which higher values assigned to strings $x \in \SigmaStar$ correspond to greater confidence that $x \in \Language$ \citep[cf.][for data and argumentation in favor of a gradient model of phonotactic acceptability in humans]{albright2009feature}. 

We may then characterize the language learning problem as one of acquiring a collection of pairs $(x_1,y_1), (x_2,y_2), \dots, (x_n,y_n)$ where $x_i \in \SigmaStar$, and $y_i \in \{0,1\}$ correspond to \textbf{acceptability judgments} about whether $x_i \in \Language$. Given this data, a learner's job is to identify a language consistent with these pairs. Importantly, in this setting, learners may have access to both positive and negative evidence. 

\paragraph{Approach}

In our problem characterization, the data acquisition process takes place over a series of time steps. At each time step $t$, the learner uses a \key{policy} $\policy$ according to which a new string 
$x_t \in \candpoolrandommath$ is selected; here \candpoolrandom is some set of possible strings, with $\Language \subset \candpoolrandommath \subset \SigmaStar$.
The chosen string is then passed to an \key{informant} that provides the learner a value $y_t \in \{0,1\}$ corresponding to whether $x_t$ is in $\Language$. The new datum $(x_t,y_t)$ is then appended to a running collection of (string, judgment) pairs $(\Xmath,\Ymath)$, after which the learning process proceeds to the next time step. This procedure is summarized in \Cref{alg:query}.

Conceptually, there are two ways in which a learner might gather information about a new language. One possibility is to gather examples well-formed strings already produced by users of the language (e.g.\ by listening to a conversation, or collecting a text corpus), similar to an \say{immersion} approach when learning a new language.
In this case, the learner does not have control over the specific selected string $x_t$, but it is guaranteed that the selected string is part of the language: $x_t \in \Language$ and thus $y_t=1$. 

The other way of collecting information is to select some string $x_t$ from \candpoolrandom, and directly elicit a judgment $y_i$ from a knowledgeable informant. This approach is often pursued by linguists working with language informants in a documentation setting, where their query stems from a hypothesis about the structural principles of the language. Here, examples can be chosen to be maximally informative, and negative evidence gathered directly.
In practice, learners might also use \say{hybrid policies} that compare which of multiple basic policies (passive observation, active inquiry) is expected to yield a new datum that optimally improves the learner's knowledge state. %
Each of these strategies is described in more detail below.

\begin{algorithm}[t!]
    \caption{Iterative Query Procedure}\label{alg:query}
    \KwIn{policy $\pi$, total timesteps $T$}
    $(\Xmath, \Ymath) \gets [\;]$; $t \gets 0$\;
    \While{$t < T$}{
        $x_t \gets \pi(x \mid \Xmath,\Ymath)$\;
        $y_t \gets \text{informant}(x_t)$\;
        append $(x_t, y_t)$ to $(\Xmath, \Ymath)$\;
        $t \gets t + 1$\;
    }
\end{algorithm}

\paragraph{Model assumptions}
\label{sec:model-assumptions}

To characterize the learning policies, we make the following assumptions regarding the \key{model} trained from available data $(\Xmath,\Ymath)$. We assume that the function $f : \SigmaStar \to [0,1]$ acquired from $(\Xmath, \Ymath)$ can be interpreted as a conditional probability of the form $p(y\mid x, \Xmath,\Ymath)$. We further assume that this conditional probability is determined by a set of parameters $\fullthetamath$ for which a(n approximate) posterior distribution $P(\fullthetamath \mid \Xmath, \Ymath)$ is maintained, with $p(y\mid x, \Xmath,\Ymath) = \int_{\fullthetamath}  P(y|x,\fullthetamath) P(\fullthetamath \mid \Xmath, \Ymath)\, d\fullthetamath$.

\section{Query policies}
\label{sec:strategies}
In the framework described in \cref{sec:problem}, how should a learner choose which questions to ask the informant?
Below, we describe a family of different policies for learning.

\subsection{Basic policies}
\label{sec:basic-policies}

\paragraph{Train} 
\label{par:train}
The first basic policy, $\atrainmath(x\mid {\Xmath, \Ymath})$, corresponds to observing and recording an utterance by a speaker. For simplicity we model this as uniform sampling (without replacement) over $\Language$:
\begin{align*}
\atrainmath(\policyinputmath) \sim U(\{x \in \Language - \Xmath \}).
\end{align*}

\paragraph{Uniform} 
\label{par:unif}
The second basic policy, $\aunifmath(\policyinputmath)$, samples a string uniformly from \candpoolrandom 
and presents it to the informant for an acceptability judgment:
\begin{align*}
\aunifmath(\policyinputmath) \sim U(\{ x \in \candpoolrandommath \}).
\end{align*}

\paragraph{Label Entropy} 
\label{par:label-entropy}
The $\alabelentmath(\policyinputmath)$ policy selects the string $\SelectedString$ with the maximum entropy \entropy over labels $y$ under the current model state:
\begin{align*}
\SelectedString= & \argmax_{x \in \candpoolrandommath} \entropymath(y\mid x,\Xmath,\Ymath).
\end{align*}

\paragraph{Expected Information Gain}
\label{par:info-gain}
The $\ainfomath(\policyinputmath)$ policy selects the candidate that, if observed, would yield the greatest expected reduction in entropy
over the posterior distribution of the model parameters \fulltheta. This is often called the \textbf{information gain} \citep{mackay-1992}; we denote the change in entropy as $\metricinfogainmath(x,y;\Xmath,\Ymath)$:
\begin{equation}
\begin{aligned}
\label{eq:v-info}
\metricinfogainmathbold&(x, y;\Xmath,\Ymath)\\
&=\entropymath(\fullthetamath\mid  \Xmath,\Ymath) - \entropymath(\fullthetamath\mid  x, y, \Xmath,\Ymath).
\end{aligned}
\end{equation}

The expected information gain policy selects the $x^*$ that maximizes $\mathop{\mathbb{E}}_{y \in [0, 1]} \metricinfogainmath(x,y;\Xmath,\Ymath)$, \ie:
\begin{align*}
x^* = &\argmax_{x \in \candpoolrandommath}\\ 
& \quad\quad \metricinfogainmath(x,y=1;\Xmath,\Ymath) \cdot p(y=1\mid x, \Xmath,\Ymath) \\
& ~~~ +  \metricinfogainmath(x,y=0;\Xmath,\Ymath) \cdot p(y=0\mid x, \Xmath,\Ymath),\\
\ainfomath&(\policyinputmath)=\delta(x^*),
\end{align*}
where $\delta(x)$ denotes the probability distribution that places all its mass on $x$.

\subsection{Hybrid Policies}
\label{s:hybrid_policies}

\textit{Hybrid} policies dynamically choose at each time step among a set of basic policies $\Pi$ based on some metric $V$. %
At each step, the hybrid policy estimates the expected value of $V$ for each basic policy $\pi \in \Pi$, chooses the policy $\pi*$ that has the highest expected value, and then samples $x \in \SigmaStar$ according to $\pi^*$. Here, we study one such policy:
    $\Pi = [\atrainmath, \ainfomath]$, with metric $V=\metricinfogainmath$.
We refer to the non-train policy as \nontrain and the metric used to select $\pi^* \in [\nontrainmath, \atrainmath]$ at each step as $V$.%

We explore two general methods for estimating the expected value of $V$ for each policy $\pi^*$: \textit{history}-based and \textit{model}-based. We also explore a \textit{mixed} approach using a history-based method for $\atrainmath$ and a model-based method for \nontrain.

\paragraph{History} 
\label{par:history}
In the {history-based} approach, the model keeps a running average of empirical values of $V$ for candidates previously selected by \atrain and \nontrain.

For instance, for history-based hybrid policy $\infogaintrainhistorymath(\policyinputmath)$, $V=\metricinfogainmath$ (see Table~\ref{subtab:hybrid_policies}). Suppose at a particular step, the basic policy $\pi^*$ selected by \infotrainhistory chose query $x$, which received label $y$ from the informant. Then the history-based \infotrainhistory would store the empirical information gain between
$p(\fullthetamath\mid \Xmath,\Ymath), p(\fullthetamath\mid x,y,\Xmath,\Ymath)$ for the chosen $\pi^*$; in future steps, it would then select the $\pi^* \in [\atrainmath, \nontrainmath]$ with the highest empirical mean of $V$, in this case the empirical mean information gain, over candidates queried by each basic policy.

\hypertarget{score:empirical}{}
More formally, let $\scoreempiricalmathbold(\pi;\Xmath,\Ymath)$ refer to the mean of observed values $V$ for candidates $x_i$ selected by $\pi$ before step $t$, where $\pi \in [\atrainmath, \nontrainmath]$: 
\begin{align*}    
\scoreempiricalmathbold(\pi;\Xmath,\Ymath)&=\frac{\sum_{i \in I_{\pi}} V(x_i, y_i;\Xmath_{<i})}{|I_{\pi}|},\\
    \text{where } I_{\pi}&=\{i \mid x_i \text{ was selected by } \pi, i < t\}.
\end{align*}
$V(x_i, y_i;\Xmath_{<i})$ denotes $V$'s score for the \textit{i}'th data-point $x_i$ selected by $\pi$ under a model that as observed data $\Xmath_{<i}, \Ymath_{<i}$.

Then at step $t$, the history-based hybrid policies sample $\pi^*$ according to:
\begin{align*}
\pi^* &= \argmax_{\pi \in [\nontrainmath, \atrainmath]} \scoreempiricalmath(\pi;\Xmath,\Ymath).
\end{align*}

For $t < 2$, we automatically select \atrain and \nontrain in a random order, each once, to ensure we have empirical means for both policies.

\paragraph{Model} 
\label{par:model}
The {model-based} approach is prospective and involves using the current posterior distribution over model parameters to compute an expected value for the target metric under the policy. We define two ways of computing these expectations. 

\hypertarget{score:expectation-nontrain}{}
$\scoremodelnontrainmath$ computes an expectation over possible \textbf{labels} $y$ for the candidate $x^*$ that will be chosen by policy $\pi$. We use $\scoremodelnontrainmath$ to score \textbf{non-train} basic policies \nontrain because they select $x^*$ deterministically given \candpoolrandom, \ie selecting the inputs that maximize the objectives described in \sect{sec:basic-policies}. More formally:
\begin{align*}
\scoremodelnontrainmathbold(\nontrainmath;\Xmath,\Ymath)&= \mathop{\mathbb{E}}_{y \in [0, 1]} V(x^*, y;\Xmath,\Ymath), x^* \sim \nontrainmath.
\end{align*}

\hypertarget{score:expectation-train}{}
$\scoremodeltrainmath$ computes an expectation over possible \textbf{inputs} $x \in \candpooltrainmath$ and assumes a fixed label $(y=1)$. We score the \textbf{train} basic policy \atrain with $\scoremodeltrainmath$ because the randomness for \atrain is over forms in the lexicon that could be sampled by \atrain, and labels are always 1. More formally:
\begin{align*}
\scoremodeltrainmathbold(\atrainmath;\Xmath,\Ymath)&= \mathop{\mathbb{E}}_{x \in \candpooltrainmath} V(x, y=1;\Xmath,\Ymath).
\end{align*}
In practice, however, we approximate this expectation with samples from \candpoolrandom,  since we do not assume that the model has access to the lexicon used by the informant. In particular, we model the probability that a form $x$ is in the lexicon as $p(y=1\mid x;\Xmath,\Ymath)$.

Using the policy-specific expectations defined above, the model-based approach selects the policy $\pi^*$ according to:
\begin{align*}
\pi^* = \argmax_{\pi \in [\nontrainmath, \atrainmath]} S(\pi;\Xmath,\Ymath).
\end{align*}

\paragraph{Mixed} 
\label{par:mixed}
Finally, the {mixed} policies combine the retrospective evaluation of the history-based method and the prospective evaluation of the model-based method.
In particular, we use the \textbf{model}-based approach for non-train \nontrain (\ie scoring with \hyperlink{score:expectation-nontrain}{$\scoremodelnontrainmath$}) and the \textbf{history}-based approach for \textbf{train} policy \atrain (\ie scoring with \hyperlink{score:empirical}{$\scoreempiricalmath$}):
\begin{align*}
S(\nontrainmath;\Xmath,\Ymath)&=\scoremodelnontrainmath(\nontrainmath;\Xmath,\Ymath),\\
S(\atrainmath;\Xmath,\Ymath)&=\scoreempiricalmath(\atrainmath;\Xmath,\Ymath),\\
\pi^* &= \argmax_{\pi \in [\nontrainmath, \atrainmath]} S(\pi;\Xmath,\Ymath).
\end{align*}

For $t=0$, we always select \atrain to ensure we have an empirical mean for \atrain. \Cref{tab:summary_of_query_policies} provides an overview of the query policies described in the preceding sections.

\newcommand{\empiricalmeanlink}{\hyperlink{score:empirical}{$\scoreempiricalmath$}\xspace}

\newcommand{\expectationtrainlink}{\hyperlink{score:expectation-train}{$\scoremodeltrainmath$}\xspace}

\newcommand{\expectationnontrainlink}{\hyperlink{score:expectation-nontrain}{$\scoremodelnontrainmath$}\xspace}

\newcommand{\empiricalmean}{{$\scoreempiricalmath$}\xspace}

\newcommand{\expectationtrain}{{$\scoremodeltrainmath$}\xspace}

\newcommand{\expectationnontrain}{{$\scoremodelnontrainmath$}\xspace}

\newcommand{\mixed}{\hyperref[par:mixed]{Mixed}\xspace}
\newcommand{\modellink}{\hyperref[par:model]{Model}\xspace}
\newcommand{\history}{\hyperref[par:history]{History}\xspace}

\begin{table*}[!htbp]
\centering
\begin{subtable}[t]{0.22\linewidth}
    \centering
    \footnotesize
    \begin{tabular}{@{}ll@{}}
        \toprule
        \multirow{2}{*}{\begin{tabular}{@{}c@{}}Basic \\ Policy \end{tabular}} & \multirow{2}{*}{\begin{tabular}{@{}c@{}}Quantity \\ Maximized \end{tabular}} \\\\
        \midrule
        \hyperref[par:train]{\atrain} & -- \\
        \hyperref[par:unif]{\aunif} & -- \\
        \hyperref[par:label-entropy]{\alabelent} & {Label entropy} \\
        \hyperref[par:info-gain]{\ainfo} & Expected info gain \\
        \bottomrule
    \end{tabular}
    \caption{
    Basic policies (\sect{sec:basic-policies}).
    }
    \label{subtab:basic_policies}
\end{subtable}%
\hspace{0.05\linewidth}%
\begin{subtable}[t]{0.63\linewidth}
    \centering
    \footnotesize
    \begin{tabular}{@{}lc|cccc@{}}
        \toprule

        \multicolumn{1}{c}{Hybrid} & \multicolumn{1}{c}{Basic} & \multicolumn{4}{|c}{Basic Policy Selection} \\

        \multicolumn{1}{c}{Policy} & \multicolumn{1}{c}{Choices $\Pi$} & \multicolumn{1}{|c}{Method} & \multicolumn{1}{c}{Metric $V$} & \atrain score & Non-train score \\
        
        \midrule

        \infotrainhistory & \multirow{3}{*}{\atrain, \ainfo} & \history & \multirow{3}{*}{\begin{tabular}{@{}c@{}}Info gain \\ (\metricinfogain, Eq~\ref{eq:v-info})\end{tabular}} & \empiricalmean & \empiricalmean\\
        
        \infotrainmodel && \modellink && \expectationtrain & \expectationnontrain \\
        
        \infotrainmixed && \mixed && \empiricalmean & \expectationnontrain \\
        \bottomrule
    \end{tabular}
    \caption{
    Hybrid policies (\sect{s:hybrid_policies}).
    }
    \label{subtab:hybrid_policies}
\end{subtable}
\caption{
Summary of query policies (\sect{sec:strategies}). \empiricalmeanlink refers to empirical mean. \expectationnontrainlink and \expectationtrainlink refer to the expectation metrics for the non-train \nontrain and train \atrain strategies, respectively. \todo{Clean the strategy selection notation/make clearer} \todo{Need better term than "Quantity Maximized" (but this shouldn't be metric, bc not the same as metric in the hybrid case)}
\todo{Confirm links: link train to train-unif sentence or train paragraph? tweak links to the metrics V?}
 Basic policies select inputs to query the informant. 
Hybrid policies choose between a set of basic policies $\Pi$ by scoring them with a metric $V$ and one of the scoring functions.}
\label{tab:summary_of_query_policies}
\vspace{-.5em}
\end{table*}

\usetikzlibrary{bayesnet}
\usetikzlibrary{arrows}

\section{A Grammatical Model for Phonotactics}
\label{s:learner_structure}

We implement and test our approach for a simple categorical model of phonotactics.
The grammar consists of two components. First, a finite set of phonological feature functions $\{\phi_i\}:\SigmaStar \mapsto \{0,1\}$; if $\phi_i(x)=1$ we say that feature $i$ is \key{active} for string $x$.   This set is taken to be universal and available to the learner before any data are observed.
 Second, a set of binary values $\fullthetamath=\{\theta_i\}$, one for each feature function; if $\theta_i=1$ then feature $i$ is \key{penalized}. In our simple categorical model, a string is grammatical if and only if no feature active for it is penalized. $\fullthetamath$ thus determines the language: $L = \{x : \sum_i \theta_i(x) \phi_i(x) = 0 \}$.
We assume a factorizable prior distribution over which features are active:
 $p(\fullthetamath)=\prod_{\theta_j\in\fullthetamath}p(\theta_j)$. To enable mathematical tractability, we also incorporate a noise term $\alpha$ which causes the learner to perceive a judgment from the informant as noisy (reversed) with probability $1-\alpha$.%

This model is based on a decades-long research tradition in theoretical and experimental phonology into what determines the range and frequency of possible word forms in a language. A consensus view of the topic is that speakers have fine-grained judgments about the acceptability of nonwords (for example, most speakers judge \textit{blick} to be more acceptable than \textit{bnick}; \citealp{chomsky1968sound}), and that this knowledge can be decomposed into the independent, additive effects of multiple prohibitions on specific sequences of sounds (in phonological theory, termed \textsc{Markedness} constraints). Further, speakers form these generalizations at the level of the phonological feature, since they exhibit structured judgments that distinguish between different unattested forms: speakers systematically rate \textit{bnick} as less English-like than \textit{bzick}, despite no attested words having initial \textit{bn-} or \textit{bz-}. We reflect this knowledge in our generative model: to determine the distribution of licit strings in a language, we first sample some parameters which govern subsequences of features which are penalized in the language.

In our model we take $\{\phi_i\}$ to be a collection of phonological \textbf{feature trigrams}: an ordered triple of three phonological features with values that pick out some class of trigrams of segments in the language (see \sect{sec:atr_harmony} for more details and examples). Since our phonotactics are variants on vowel harmony, these featural trigrams are henceforth assumed to be relativized to the vowel tier, regulating vowel qualities in three adjacent syllables. %
In order to capture generalizations that may hold differently in edge-adjacent vs. word-medial position, we pad the representation of each word treated by the model with a boundary symbol \say{\#} --- omitted generally in this paper, for simplicity --- which bears the  [$+$ word boundary] feature that the trigram constraints can refer to (following the practice of \citealp{hayes2008maximum}, inspired by \citealp{chomsky1968sound}).

\subsection{Implementation details}
\label{s:implementation}
Our general approach and specific model create several computational tractability issues that we address here. First, all policies aside from $\atrainmath$ and \aunif in principle require search for an optimal string $x$ within \candpoolrandom. In practice, we consider $\candpoolrandommath = \SigmaStar\{2, 5\}$, \ie \candpoolrandom is the set of strings with 2-5 syllables. This resulting set is still very large, so we approximate the search over \candpoolrandom by uniformly sampling a set of $k$ candidates and choosing the best according to $V$.
We sample candidates by uniformly sampling a length, then uniformly sampling each syllable from the inventory of possible onset-vowel combinations in the language (with replacement). We then de-duplicate candidates and filter $\Xmath$, excluding previously observed sequences and those that were accidental duplicates with items in the test set.

Second, although the model parameters $\fullthetamath$ are independent in the prior, conditioning on data renders them conditionally dependent and computing with the true posterior is in general intractable. To deal with this, we use mean-field variational Bayes to approximate the posterior as $p(\fullthetamath \mid \Xmath, \Ymath) \approx \prod_{\theta_j\in\fullthetamath} q(\theta_j=1 \mid \Xmath, \Ymath)$. We use this approximation to both estimate the model's posterior (used by \alabelent and \ainfo) %
and to make predictions about individual new examples. See Appendix~\ref{sec:derivation} for details. %

\section{Experiments}
\label{sec:experiments}

We now describe our experiments for evaluating the different query policies. We evaluate on two types of languages. We call the first the ATR Vowel Harmony language (\sect{sec:atr_harmony}), which has  grammar that regulates the distribution of types of vowels, inspired by those found in many languages of the world. The purpose of evaluating on this language is to evaluate how well our new approach, and specifically the various non-baseline query policies, work on naturalistic data.
We also evaluate on a set of procedurally-generated languages (\sect{sec:atr_generated}) that are matched on statistics to ATR Vowel Harmony, \ie they have the same number of feature trigrams that are penalized, but differ in \emph{which}. This second set of evaluations aims to determine how robust our model is to typologically-unusual languages, so we can be confident that any success in learning ATR Vowel Harmony is attributable to our procedure, rather than a coincidence of the typologically-natural vowel harmony pattern.

These experiments lead to three sets of analyses: in the first (\sect{s:analysis-in-distrib}), we both select hyperparameters and evaluate on procedurally-generated languages through \textit{k}-fold cross validation. These results can be interpreted as an in-distribution analysis of the query policies. In the second set of results (\sect{s:analysis-out-distrib}), we evaluate the policies out-of-distribution by selecting hyperparameters on the procedurally-generated languages and evaluating on the ATR Vowel Harmony language. In the last analysis (\sect{s:analysis-exploratory}), we evaluate the upper bound of policy performance by selecting hyperparameters and evaluating on the same language, ATR Vowel Harmony.

\subsection{ATR Vowel Harmony}
\label{sec:atr_harmony}

We created a model language whose words are governed by a small set of known phonological principles. Loosely inspired by harmony systems common among Niger-Congo and Nilo-Saharan languages spoken throughout Africa, the vowels in this language can be divided into two classes, defined with respect to the phonological feature Advanced Tongue Root (ATR); for typological data, see \citet{casali2003atr, casali2008atr, casali2016some,rose2018atr}, among others. In this language, vowels that are [+ATR] are \{i, e\}, and have pronunciations that are more peripheral in the vowel space; those that are [-ATR] are \{\textsci, \textepsilon\}, and are more phonetically centralized. For the sake of simplicity, we restrict the simulated language to only have front vowels. A fifth vowel in the system, [a], is not specified for ATR. This language has consonants \{p, t, k, q\}, which are distributed freely with respect to one another and to vowels with the exception that syllables must begin with exactly consonant and must contain exactly one vowel, a typologically common restriction. Since consonants are not regulated by the grammar we are working with, the three binary features (leaving out [word boundary]) create a set of 512 possible feature trigrams which characterize the space of all possible strings in the language. The syllable counts of words follows a Poisson distribution with $\lambda$ $=$ 2. 

The single rule active in this language governs the distribution of vowels specified for the feature [ATR]: vowels in adjacent syllables had to have the same [ATR] specification. This means that vowel sequences in a word can be [i\ldots e] or [\textepsilon \ldots \textsci], but not [e \ldots \textepsilon] or [e \ldots \textsci]. Since [a] is not specified for ATR, it creates boundaries that allow different ATR values to exist on either side of it: for example, while the vowel sequence [e \ldots \textepsilon] is not permitted, the sequence [e \ldots a \ldots \textepsilon] is allowed, because the ATR-distinct vowels are separated by the unspecified [a].  This yielded sample licit words like [katipe], [t\textepsilon p\textsci], and [qekat\textsci], and illicit ones [k\textepsilon kiqa], [t\textsci taqik\textepsilon], and [qiq\textsci ka].

Feature trigrams were composed of triples of the features and specifications shown in Appendix \Cref{tab:atr_features}, any one of which picks out a certain set of vowel trigrams in adjacent syllables.

\paragraph{Data} We sampled 157 unique words as the lexicon \candpooltrain, and a set of 1,010 random words, roughly balanced for length, as a test set. The model was provided with the set of features in Appendix \Cref{tab:atr_features}, and restrictions on syllable structure 
for use in the proposal distribution. 

\paragraph{Informant} The informant was configured to reject any word that contained vowels in adjacent syllables that differed in ATR specification (like [pekit\textepsilon] or [qetat\textsci kipe]), and accept all others. 

\subsection{Procedurally-Generated Languages}
\label{sec:atr_generated}

We also experimented with languages that share the same feature space, and thus the same set of 512 feature trigrams, as ATR Vowel Harmony (\sect{sec:atr_harmony}) but were \emph{procedurally generated} by sampling 16 of the 512 total feature trigrams to be \say{on} (\ie penalized) and set all others to be off, creating languages with different restrictions on licit vowel sequences in adjacent syllables.

\paragraph{Data}
For each \say{language} (\ie set of sampled feature trigrams to be penalized), we carried out a procedure to sample the lexicon \candpooltrain, as well as evaluation datasets.
For each set of 16 $\theta$ values representing penalized phonological feature trigrams, we created random strings as in Experiment 1, filtering them to ensure that the train and test set are of equal size, and the test set is balanced for length of word and composed of half acceptable and half unacceptable forms. %

\subsection{Experimental Set-Up}
\label{s:set-up}
\paragraph{Hyperparameters} 
The model has several free parameters: a noise parameter ${\alpha}$ that represents the probability that an observed label is correct (versus noisy),
and ${\priormath}$, the prior probability of a feature being \textit{on} (penalized), \ie $p_{\text{prior}}(\theta_j=1)$. There are also hyperparameters governing the optimization of the model: we denote by $s$ the number of optimization steps in the variational update.\footnote{These optimization parameters govern both the model's learning and the evaluation of candidate queries for prospective strategies, \ie \ainfo,  and the hybrid strategies. %
} 
When $s=\infty$, we optimize until the magnitude of the change in $\fullthetamath$ is less than or equal to an error threshold $\epsilon = 2e^{-7}$  We also experiment with $s=1$, in which we perform a single update.
 
We ran a grid-search over the parameter space of log(log($\alpha)) \in$ \{0.1, 0.25, 0.5, 1, 2, 4, 8\}, $\priormath \in$ \{0.001, 0.025, 0.05, 0.1, 0.2, 0.35\}, and $s \in \{1, \infty\}$. We ran 10 random seeds (9 for the procedurally generated languages)\footnote{For the generated languages, seed also governed the ``language,'' \ie phonological feature trigrams sampled as ``on.''} and all query policies in \Cref{tab:summary_of_query_policies} for each hyperparameter setting. Each experiment was run for 150 steps.

For non-train policies, we generated $k=100$ candidates from \candpoolrandom. 

\paragraph{Evaluation} 

At each step, we compute the AUC (area under the ROC curve) on the test set. We then compute the mean AUC value across steps, which we refer to as the \textbf{mean-AUC}; a higher mean-AUC indicates more efficient learning. We report the median of the mean-AUC values over seeds.
\begin{figure*}[h!] 
\centering
 \makebox[\textwidth]{\includegraphics[width=\textwidth]{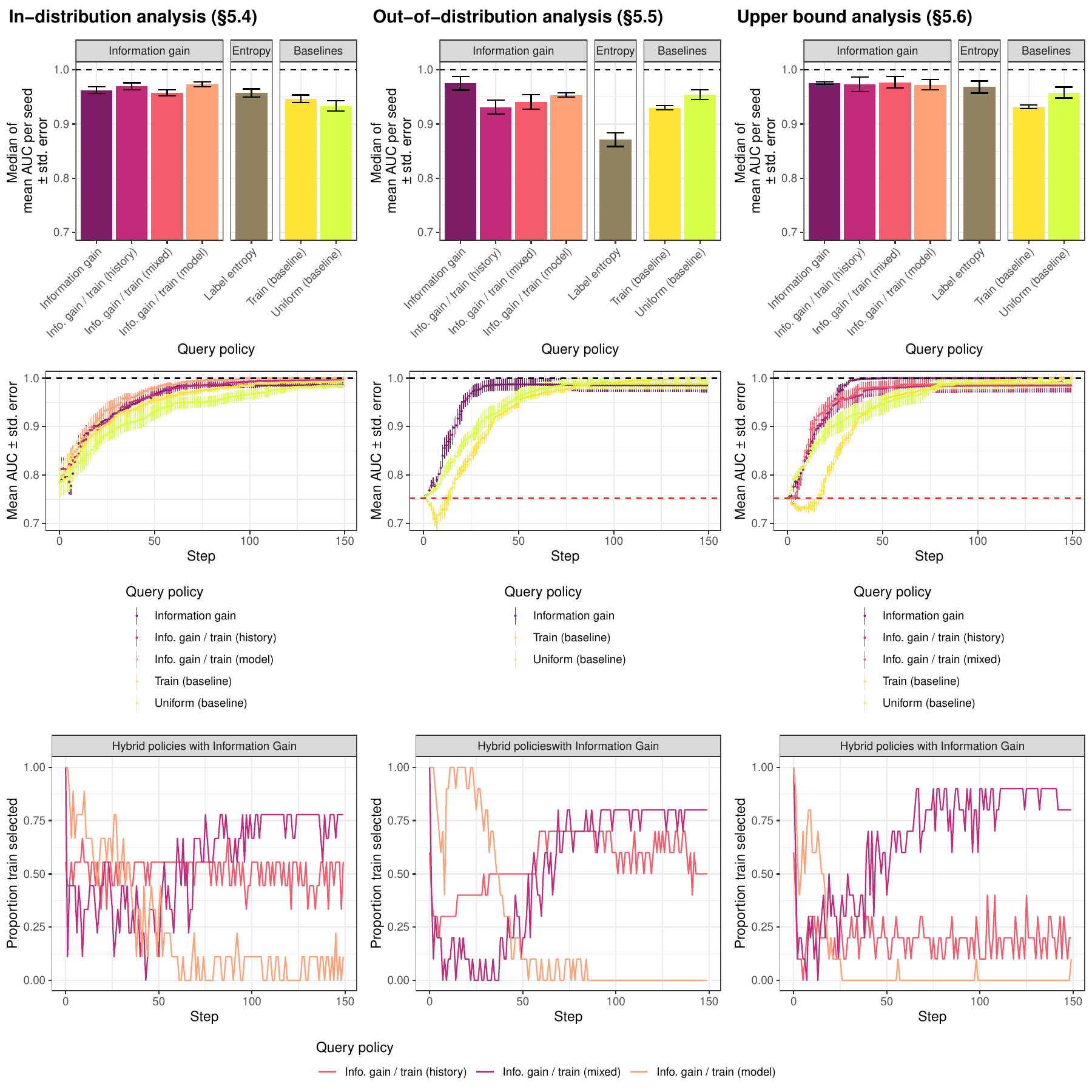}}
\caption{We report three analyses of the toy ATR Vowel Harmony language and our procedurally-generated languages: in-distribution (left column, see \sect{s:analysis-in-distrib}), out-of-distribution (center column, see \sect{s:analysis-out-distrib}), and an upper-bound assessment (right column, see \sect{s:analysis-exploratory}). For each, we report the median and standard error of the mean-AUC over steps aggregated across runs (top row; numerical values and hyperparameters reported in Appendix \Cref{tab:hyperparams}), average AUC at each step aggregated across runs (middle row), and at each step the proportion of runs where the basic \textit{train} strategy was selected by the hybrid strategies (bottom row). \textbf{Results:} In terms of median mean-AUC (top row), our query strategies are numerically on par with, if not beating, the stronger of the two baseline conditions; statistically, only the difference between \textit{Info.~gain / train (model)} and \textit{uniform} was significant in the in-distribution analysis (top left). Average AUC over time (middle row) shows a similar pattern across all three analyses, with the non-baseline strategies rising faster and asymptoting sooner than baseline strategies, but usually with greater variance. Finally, though all hybrid strategies prefer non-train some portion of the time, the \textit{Info.~gain / train (model)} exhibits an interpretable shift from early preference for \textit{train} data to later preference for its own synthesized queries in all three analyses.}
\label{fig:full_page_figure}
\end{figure*}

\subsection{In-Distribution Results}
\label{s:analysis-in-distrib}
Assessing the in-distribution results, shown in the left column of \Cref{fig:full_page_figure}, we see that interactive elicitation is on par with, if not higher than, baseline strategies (top left plot). The difference between the \textit{train} and \textit{uniform} baselines was not significant according to a two-sided paired \textit{t}-test, and the only strategy that performed significantly better than \textit{train} after correcting for multiple comparisons was \textit{Info.~gain / train (model)}. This difference is more visually striking in the plot of average AUC over time (middle left plot), where \textit{Info.~gain / train (model)} both ascends faster, and asymptotes earlier, than \textit{train}, although with greater variance across runs. In the bottom left plot of \Cref{fig:full_page_figure}, we see that %
the numerically-best-performing \textit{Info.~gain / train (model)} strategy moves rather smoothly from an initial \textit{train} preference to an \textit{Info.~gain} preference as learning progresses. That is, information in known-good words is initially helpful, but quickly becomes less useful as the model learns more of the language and can generate more targeted queries. %

\subsection{Out-Of-Distribution Results}
\label{s:analysis-out-distrib}
The out-of-distribution analysis on the ATR Vowel Harmony language found greater variance of median mean-AUC between strategies, and also greater variance within strategies across seeds (top center plot). We note that this performance is lower than what is found in the upper-bound analysis, since the hyperparameters (listed in Appendix \Cref{tab:hyperparams}) were chosen based on the pooled results of the procedurally-generated languages. As in the in-distribution analysis, we found no statistical difference between the two baselines, nor between the \textit{Info.~gain} strategy and \textit{uniform}, although \textit{Info.~gain} performed numerically better. In terms of average AUC over time (middle center plot), we find again that the top two non-baseline strategies rise faster and peak earlier than \textit{uniform}, but exhibit greater variance.

\subsection{Upper Bound Results}
\label{s:analysis-exploratory}

Greedily selecting for the best test performance in a hyperparameter search conducted on ATR Vowel Harmony yields superior performance compared to the out-of-distribution analysis hyperparameters, as seen in the top right plot in \Cref{fig:full_page_figure}. Appendix \Cref{tab:hyperparams} lists the hyperparameter values used. However, we found no significant difference between the stronger baseline (\textit{uniform}) and any other strategy after correcting for multiple comparisons.

\section{Related Work}

The goal of \textbf{active learning} is to improve learning efficiency by allowing models to choose which data to query an oracle about \citep{zhang-etal-2022-survey}. \emph{Uncertainty sampling} \citep{uncertainty-sampling} methods select queries for which model uncertainty is highest. Most closely related are uncertainty sampling methods for probabilistic models, including least-confidence \cite{culotta-mccallum}, margin sampling \cite{scheffer-2001}, and entropy-based methods. 

\emph{Disagreement-based strategies} query instances that maximize disagreement among a group of models \citep{seung-1992}. The distribution over a single model's parameters can also be treated as this ``group'' of distinct models, as has been done for neural models \citep{bald}.  Such methods are closely related to the feature entropy querying policy that we explore.

Another class of \emph{forward-looking methods} incorporates information about how models would change if a given data-point were observed. Previous work includes methods that sample instances based on expected loss reduction \citep{Roy2001TowardOA}, expected information gain \citep{mackay-1992}, and expected gradient length \citep{settles-2007}. These methods are closely related to the policies based on information-gain that we explore.

Our hybrid policies are also related to previous work on dynamic selection between multiple active learning policies, such as DUAL \citep{dual}, which dynamically switches between density and uncertainty-based strategies. 

The model we propose is also related to a body of work in computational and theoretical linguistics focused on \textbf{phonotactic learning}. Much of this work, largely inspired by \citet{hayes2008maximum}, seeks to discover and/or parameterize models of phonotactic acceptability on the basis of only positive data, in line with common assumptions about infant language acquisition \cite{albright2009feature,adriaans2010adding,linzen2015model,futrell2017generative,mirea2019using,gouskova2020inducing,mayer2020phonotactic,dai2023rethinking,daitoappearexception}. Our work differs from these in that we are explicitly not seeking to model phonotactic learning from the infants' point of view, instead drawing inspiration from the strategy of a linguist working with a competent native speaker to discover linguistic structure via iterated querying. Practically, this means that our model can make use of both positive and negative data, and also takes an active role in seeking out the data it will learn from.

\section{Conclusion}
\label{sec:conclusion}

We have described a method for parameterizing a formal model of language via efficient, iterative querying of a black box agent. We demonstrated that on an in-distribution set of toy languages, our query policies consistently outperform baselines numerically, including a statistically-significant improvement for the most effective policy. The model struggles more on out-of-distribution languages, though in all cases the query policies are numerically comparable to the best baseline. We note that a contributing factor to the difficulty of the query policies consistently achieving a \textit{significantly} higher performance than baselines is the small number of seeds, which exhibit nontrivial variance, particularly in hybrid policies. Future work may address this with more robust experiments.

\section*{Acknowledgements}
Thanks to members of the audience at Interactions between Formal and Computational Linguistics (IFLG) Seminar hosted by the \textit{Linguistique Informatique, Formelle et de Terrain }group, as well as two anonymous SCiL reviewers, for helpful questions and comments.

We acknowledge the following funding sources: MIT-IBM Watson AI Lab (CB, RL), NSF GRFP grant number 2023357727 (AR), a MIT Shillman Fellowship (AR), a MIT Dean of Science Fellowship (AM), and NSF IIS-2212310 (JA, AR).

\bibliography{anthology,custom}

\begin{thebibliography}{28}
\expandafter\ifx\csname natexlab\endcsname\relax\def\natexlab#1{#1}\fi

\bibitem[{Adriaans and Kager(2010)}]{adriaans2010adding}
Frans Adriaans and Ren{\'e} Kager. 2010.
\newblock Adding generalization to statistical learning: The induction of phonotactics from continuous speech.
\newblock \emph{Journal of Memory and Language}, 62(3):311--331.

\bibitem[{Albright(2009)}]{albright2009feature}
Adam Albright. 2009.
\newblock Feature-based generalisation as a source of gradient acceptability.
\newblock \emph{Phonology}, 26(1):9--41.

\bibitem[{Bender et~al.(2021)Bender, Gebru, McMillan-Major, and Shmitchell}]{bender2021dangers}
Emily~M Bender, Timnit Gebru, Angelina McMillan-Major, and Shmargaret Shmitchell. 2021.
\newblock On the dangers of stochastic parrots: Can language models be too big?
\newblock In \emph{Proceedings of the 2021 ACM conference on fairness, accountability, and transparency}, pages 610--623.

\bibitem[{Casali(2003)}]{casali2003atr}
Roderic~F Casali. 2003.
\newblock [atr] value asymmetries and underlying vowel inventory structure in niger-congo and nilo-saharan.

\bibitem[{Casali(2008)}]{casali2008atr}
Roderic~F Casali. 2008.
\newblock Atr harmony in african languages.
\newblock \emph{Language and linguistics compass}, 2(3):496--549.

\bibitem[{Casali(2016)}]{casali2016some}
Roderic~F Casali. 2016.
\newblock Some inventory-related asymetries in the patterning of tongue root harmony systems.
\newblock \emph{Studies in African Linguistics}, pages 96--99.

\bibitem[{Chomsky and Halle(1968)}]{chomsky1968sound}
Noam Chomsky and Morris Halle. 1968.
\newblock \emph{The sound pattern of {E}nglish}.
\newblock Harper \& Row New York.

\bibitem[{Culotta and McCallum(2005)}]{culotta-mccallum}
Aron Culotta and Andrew McCallum. 2005.
\newblock Reducing labeling effort for structured prediction tasks.
\newblock In \emph{Proceedings of the 20th National Conference on Artificial Intelligence - Volume 2}, AAAI'05, page 746–751. AAAI Press.

\bibitem[{Dai(to appear)}]{daitoappearexception}
Huteng Dai. to appear.
\newblock An exception-filtering approach to phonotactic learning.
\newblock \emph{Phonology}.

\bibitem[{Dai et~al.(2023)Dai, Mayer, and Futrell}]{dai2023rethinking}
Huteng Dai, Connor Mayer, and Richard Futrell. 2023.
\newblock Rethinking representations: A log-bilinear model of phonotactics.
\newblock \emph{Proceedings of the Society for Computation in Linguistics}, 6(1):259--268.

\bibitem[{Donmez et~al.(2007)Donmez, Carbonell, and Bennett}]{dual}
Pinar Donmez, Jaime~G. Carbonell, and Paul~N. Bennett. 2007.
\newblock \href {https://doi.org/10.1007/978-3-540-74958-5_14} {Dual strategy active learning}.
\newblock In \emph{Proceedings of the 18th European Conference on Machine Learning}, ECML '07, page 116–127, Berlin, Heidelberg. Springer-Verlag.

\bibitem[{Futrell et~al.(2017)Futrell, Albright, Graff, and O’Donnell}]{futrell2017generative}
Richard Futrell, Adam Albright, Peter Graff, and Timothy~J O’Donnell. 2017.
\newblock A generative model of phonotactics.
\newblock \emph{Transactions of the Association for Computational Linguistics}, 5:73--86.

\bibitem[{Gal et~al.(2017)Gal, Islam, and Ghahramani}]{bald}
Yarin Gal, Riashat Islam, and Zoubin Ghahramani. 2017.
\newblock Deep bayesian active learning with image data.
\newblock In \emph{Proceedings of the 34th International Conference on Machine Learning - Volume 70}, ICML'17, page 1183–1192. JMLR.org.

\bibitem[{Gouskova and Gallagher(2020)}]{gouskova2020inducing}
Maria Gouskova and Gillian Gallagher. 2020.
\newblock Inducing nonlocal constraints from baseline phonotactics.
\newblock \emph{Natural Language \& Linguistic Theory}, 38:77--116.

\bibitem[{Hayes and Wilson(2008)}]{hayes2008maximum}
Bruce Hayes and Colin Wilson. 2008.
\newblock A maximum entropy model of phonotactics and phonotactic learning.
\newblock \emph{Linguistic inquiry}, 39(3):379--440.

\bibitem[{Hewitt and Manning(2019)}]{hewitt2019structural}
John Hewitt and Christopher~D Manning. 2019.
\newblock A structural probe for finding syntax in word representations.
\newblock In \emph{Proceedings of the 2019 Conference of the North American Chapter of the Association for Computational Linguistics: Human Language Technologies, Volume 1 (Long and Short Papers)}, pages 4129--4138.

\bibitem[{Lewis and Gale(1994)}]{uncertainty-sampling}
David~D. Lewis and William~A. Gale. 1994.
\newblock A sequential algorithm for training text classifiers.
\newblock In \emph{SIGIR '94}, pages 3--12, London. Springer London.

\bibitem[{Linzen and O’Donnell(2015)}]{linzen2015model}
Tal Linzen and Timothy O’Donnell. 2015.
\newblock A model of rapid phonotactic generalization.
\newblock In \emph{Proceedings of the 2015 Conference on Empirical Methods in Natural Language Processing}, pages 1126--1131.

\bibitem[{MacKay(1992)}]{mackay-1992}
David J.~C. MacKay. 1992.
\newblock \href {https://doi.org/10.1162/neco.1992.4.4.590} {{Information-Based Objective Functions for Active Data Selection}}.
\newblock \emph{Neural Computation}, 4(4):590--604.

\bibitem[{Mayer and Nelson(2020)}]{mayer2020phonotactic}
Connor Mayer and Max Nelson. 2020.
\newblock Phonotactic learning with neural language models.
\newblock \emph{Society for Computation in Linguistics}, 3(1).

\bibitem[{Mirea and Bicknell(2019)}]{mirea2019using}
Nicole Mirea and Klinton Bicknell. 2019.
\newblock Using lstms to assess the obligatoriness of phonological distinctive features for phonotactic learning.
\newblock In \emph{Proceedings of the 57th annual meeting of the association for computational linguistics}, pages 1595--1605.

\bibitem[{Mohamed et~al.(2022)Mohamed, Lee, Borgholt, Havtorn, Edin, Igel, Kirchhoff, Li, Livescu, Maal{\o}e et~al.}]{mohamed2022self}
Abdelrahman Mohamed, Hung-yi Lee, Lasse Borgholt, Jakob~D Havtorn, Joakim Edin, Christian Igel, Katrin Kirchhoff, Shang-Wen Li, Karen Livescu, Lars Maal{\o}e, et~al. 2022.
\newblock Self-supervised speech representation learning: A review.
\newblock \emph{IEEE Journal of Selected Topics in Signal Processing}.

\bibitem[{Rose(2018)}]{rose2018atr}
Sharon Rose. 2018.
\newblock Atr vowel harmony: new patterns and diagnostics.
\newblock In \emph{Proceedings of the Annual Meetings on Phonology}, volume~5.

\bibitem[{Roy and McCallum(2001)}]{Roy2001TowardOA}
Nicholas Roy and Andrew McCallum. 2001.
\newblock \href {https://api.semanticscholar.org/CorpusID:14293159} {Toward optimal active learning through monte carlo estimation of error reduction}.
\newblock In \emph{International Conference on Machine Learning}.

\bibitem[{Scheffer et~al.(2001)Scheffer, Decomain, and Wrobel}]{scheffer-2001}
Tobias Scheffer, Christian Decomain, and Stefan Wrobel. 2001.
\newblock Active hidden markov models for information extraction.
\newblock In \emph{Advances in Intelligent Data Analysis}, pages 309--318, Berlin, Heidelberg. Springer Berlin Heidelberg.

\bibitem[{Settles et~al.(2007)Settles, Craven, and Ray}]{settles-2007}
Burr Settles, Mark Craven, and Soumya Ray. 2007.
\newblock \href {https://proceedings.neurips.cc/paper_files/paper/2007/file/a1519de5b5d44b31a01de013b9b51a80-Paper.pdf} {Multiple-instance active learning}.
\newblock In \emph{Advances in Neural Information Processing Systems}, volume~20. Curran Associates, Inc.

\bibitem[{Seung et~al.(1992)Seung, Opper, and Sompolinsky}]{seung-1992}
H.~S. Seung, M.~Opper, and H.~Sompolinsky. 1992.
\newblock \href {https://doi.org/10.1145/130385.130417} {Query by committee}.
\newblock In \emph{Proceedings of the Fifth Annual Workshop on Computational Learning Theory}, COLT '92, page 287–294, New York, NY, USA. Association for Computing Machinery.

\bibitem[{Zhang et~al.(2022)Zhang, Strubell, and Hovy}]{zhang-etal-2022-survey}
Zhisong Zhang, Emma Strubell, and Eduard Hovy. 2022.
\newblock \href {https://doi.org/10.18653/v1/2022.emnlp-main.414} {A survey of active learning for natural language processing}.
\newblock In \emph{Proceedings of the 2022 Conference on Empirical Methods in Natural Language Processing}, pages 6166--6190, Abu Dhabi, United Arab Emirates. Association for Computational Linguistics.

\end{thebibliography}
\bibliographystyle{acl_natbib}

\appendix

\begin{table*}[tpbh!]
\centering
\resizebox{\textwidth}{!}{%
\begin{tabular}{@{}llllll|l|llllll@{}}
\toprule
\multicolumn{6}{l|}{Out-of-distribution analysis} &
   &
  \multicolumn{6}{l}{Upper-bound analysis} \\ \midrule
\multicolumn{1}{l|}{Policy} &
  log(log($\alpha$)) &
  prior &
  \textit{s} &
  \begin{tabular}[c]{@{}l@{}}Median \\ mean-AUC\end{tabular} &
  Std. err. &
   &
  \multicolumn{1}{l|}{Policy} &
  log(log($\alpha$)) &
  prior &
  \textit{s} &
  \begin{tabular}[c]{@{}l@{}}Median \\ mean-AUC\end{tabular} &
  Std. err. \\ \cmidrule(r){1-6} \cmidrule(l){8-13} 
\multicolumn{1}{l|}{Info. gain / train   (model)} &
  0.5 &
  0.1 &
  $\infty$ &
  0.973 &
  0.004 &
   &
  \multicolumn{1}{l|}{Info. gain / train   (mixed)} &
  0.25 &
  0.1 &
  $\infty$ &
  0.977 &
  0.010 \\
\multicolumn{1}{l|}{Info. gain / train (history)} &
  1 &
  0.1 &
  $\infty$ &
  0.970 &
  0.006 &
   &
  \multicolumn{1}{l|}{Information gain} &
  0.1 &
  0.025 &
  $\infty$ &
  0.975 &
  0.002 \\
\multicolumn{1}{l|}{Info. gain / train (mixed)} &
  2 &
  0.2 &
  $\infty$ &
  0.969 &
  0.005 &
   &
  \multicolumn{1}{l|}{Info. gain / train (history)} &
  0.1 &
  0.05 &
  $\infty$ &
  0.974 &
  0.013 \\
\multicolumn{1}{l|}{Information gain} &
  0.25 &
  0.025 &
  $\infty$ &
  0.966 &
  0.004 &
   &
  \multicolumn{1}{l|}{Info. gain / train (model)} &
  1 &
  0.001 &
  1 &
  0.973 &
  0.009 \\
\multicolumn{1}{l|}{Label entropy} &
  0.1 &
  0.1 &
  $\infty$ &
  0.964 &
  0.009 &
   &
  \multicolumn{1}{l|}{Label entropy} &
  0.5 &
  0.05 &
  1 &
  0.968 &
  0.011 \\
\multicolumn{1}{l|}{Train (baseline)} &
  1 &
  0.1 &
  $\infty$ &
  0.947 &
  0.007 &
   &
  \multicolumn{1}{l|}{Uniform (baseline)} &
  0.5 &
  0.025 &
  1 &
  0.958 &
  0.010 \\
\multicolumn{1}{l|}{Uniform (baseline)} &
  1 &
  0.1 &
  1 &
  0.940 &
  0.008 &
   &
  \multicolumn{1}{l|}{Train (baseline)} &
  8 &
  0.35 &
  1 &
  0.932 &
  0.003 \\ \bottomrule
\end{tabular}%
}
\caption{Hyperparameters for the out-of-distribution analysis (\sect{s:analysis-out-distrib}) and upper-bound analysis (\sect{s:analysis-exploratory}).}
\label{tab:hyperparams}
\end{table*}

\section{Phonological features for Toy Languages}

As described in \sect{sec:atr_harmony}, the ATR Vowel Harmony language is based on the categorization of vowels as [+ATR], [-ATR], or unspecified. The features [high] and [low] also serve to distinguish vowels in the language, but are not governed by a phonotactic. In contrast, any of the 512 logically possible trigrams of specified phonological features may be penalized for the procedurally-generated languages. \Cref{tab:atr_features} displays the phonological features for each of the vowels in the languages.

\section{Hyperparameters for out-of-distribution and upper-bound analyses}

In \sect{s:set-up}, we described the hyperparameters of our grammatical model and the process by which values were selected for the out-of-distribution analysis. These selected hyperparameter values are presented in \Cref{tab:hyperparams}.

\begin{table}[b]
    \centering
    \begin{tabular}{c|ccc}
             & [high] & [low] & [ATR] \\
         \hline
         i & $+$ & $-$ & $+$ \\
         \textsc{i} & $+$ & $-$ & $-$\\
         e & $-$ &$-$  & $+$\\
         \textepsilon & $-$  &$-$  &$-$ \\
          a & $-$ & $+$ & 0\\
    \end{tabular}
    \caption{Phonological features for vowels used in the toy languages. The feature [word boundary] is omitted for simplicity, as it has the value `$-$' for all segments.}
    \label{tab:atr_features}
\end{table}

\section{Query Policy Implementation}

We now revisit the query strategies introduced in \sect{sec:strategies} and describe how they are implemented for the model described in \sect{s:learner_structure}. In particular, under the described generative model, $p(y=1\mid x,\Xmath, \Ymath)=\prod_{j \in \phi(x)}q(\theta_j=0 \mid x,\Xmath,\Ymath)$, as described above.

Let $q_y=\prod_{j \in \phi(x)}q(\theta_j=0 \mid x,\Xmath,\Ymath)$, \ie $q_y$ is the probability of label $y=1$ for input $x$ under the variational posterior; this is equivalent to the probability of all features in $\phi(x)$ being ``off''.
Let $q_{\theta_j}=q(\theta_j=1 \mid \Xmath,\Ymath)$ indicate the probability of parameter $\theta_j$ being ``on'' (\ie penalized) under the current variational $q(\fullthetamath)$.
\todo{Is this notation fine?} \todo{check whether $\phi(x)$ returns indices ($j$) or something else}
For this model, the quantities used by the query policies in \sect{sec:strategies} are computed as follows:

\paragraph{Label Entropy}
Policy \alabelent selects $x^*$ according to:
\begin{align*}
x^* = &\argmax_{x \in \candpoolrandommath} H(q_y), \text{where} \\
H(p(y\mid x,\Xmath, \Ymath))=&-q_y \log q_y \\
& -(1-q_y) \log (1-q_y).
\end{align*}

\paragraph{Expected Information Gain}

Policy \ainfo selects $x^*$ according to:
\begin{align*}
    x^* &= \argmax_{x \in \candpoolrandommath} \metricinfogainmath(x,y=1;\Xmath,\Ymath) \cdot q_y \\
        &\phantom{\argmax_{x \in \candpoolrandommath}}+ \metricinfogainmath(x,y=0;\Xmath,\Ymath) \cdot (1-q_y),
\end{align*}
where $\metricinfogainmath$ is given by
\begin{align*}
    \metricinfogainmath(x, y;\Xmath,\Ymath)&= \sum_{j \in \mid \theta\mid } \Big( H(q(\theta_j \mid \Xmath,\Ymath)) \\
    &\phantom{=} - H(q(\theta_j\mid x,y,\Xmath,\Ymath)) \Big),
\end{align*}
and $H$ is given by
\begin{align*}
    H(q(\theta_j))&=-q_{\theta_j}\log q_{\theta_j} \\
    &\phantom{=}- (1-q_{\theta_j}) \log(1-q_{\theta_j}).
\end{align*}
\todo{confirm correct sum}
\todo{Is this notation weird bc conditioning on $x,y$ but not $\Xmath, \Ymath$? is it clear what $H(q(\theta_j\mid x,y))$ should be?}

\section{Derivation of the Update Rule}
\label{sec:derivation}
\todo{derivation needs to be updated to start from ELBO}

We want to compute the posterior $p(\theta|y,x,\alpha)$, which is intractable. Thus, we approximate it with a variational posterior, composed of binomial distributions for each $\theta_i$. We further assume that the individual dimensions of the posterior (the individual components of $\theta$) have values that are not correlated. This allows us to perform coordinate ascent on each dimension of the posterior separately; thus we express the following derivation in terms of $q(\theta_i)$, where \textit{i} is the index in the feature \textit{n}-gram vector.

The variational posterior is optimized to minimize the KL divergence between the true posterior $p(\fullthetamath|X,Y,\alpha)$ and $q(\fullthetamath)$; we do this by maximizing the ELBO.

The coordinate ascent update rule for each dimension of the posterior, that is, for each latent variable, is:
$$q(\theta_i) \propto \exp \big[\mathop{\E_{q\lnot i}} \log p(\theta_i, \theta_{\lnot i}, y, x) \big].$$

Given the generative process, we can rewrite: $$p(\theta_i, \theta_{\lnot i}, y, x)=p(\theta_i) \cdot p(\theta_{\lnot i}) \cdot p(y|x,\theta_i, \theta_{\lnot i}).$$

$\mathop{\E_{q\lnot i}} \log p(\theta_{\lnot i})$ is assumed to be constant across values of $\theta_i$ (expressing the lack of dependence between parameters), so we can rewrite the update rule as:
$$q(\theta_i) \propto \exp \big[\mathop{\E_{q\lnot i}} [\log p(\theta_i) + \log p(y|x,\theta_i, \theta_{\lnot i})] \big].$$
Further, since $\log p(\theta_i)$ is constant across values of $q\lnot i$, we can rewrite it once more:
$$q(\theta_i) \propto \exp \big[ \log p(\theta_i) + \mathop{\E_{q\lnot i}} \log p(y|x,\theta_i, \theta_{\lnot i}) \big].$$

Since our approximating distribution is binomial, we describe in turn the treatment of each of the two possible values of $\theta$. First, we derive the update rule for when the label $y$ is acceptable ($y=1$). 

{We know that there are two subsets of $q\lnot i$ cases where this can happen. In $\alpha$ proportion of them, $y$ is a correct label, which can only happen when $\theta_j=0$ for all $j \neq i \in \phi(x)$. This occurs with probability $p_{\text{all\_off}}=\prod_{j \neq i \in \phi(x)}q(\theta_j=0)$.} %
There is also, then, the $1-\alpha$ proportion of cases in which $y$ is an incorrect label, and the true judgement is unacceptable. Under this assumption, at least 1 feature is on, which occurs with probability $1-p_{\text{all\_off}}$.

We can rewrite the expectation term to get approximate probabilities for both the $\theta_i=0$ and $\theta_i=1$ cases when $y=1$:
\begin{align*}
    q(\theta_i=0) &\propto \exp \big[ \log p(\theta_i=0) \\
    + \big(p_{\text{all\_off}}& \cdot \log \alpha + (1-p_{\text{all\_off}}) \cdot \log(1-\alpha)\big)\big].
\end{align*}

If $\theta_i=1,$ we know that $\log p(y|x,\theta_i, \theta_{\lnot i})=\log (1-\alpha)$ for all $q\lnot i$, since we know that $y$ must be a noisy label. Thus:

$q(\theta_i=1) \propto \exp \big[ \log p(\theta_i=1) + \log(1-\alpha)\big]$.

We can normalize these quantities to get a proper probability distribution, i.e. we can set $q(\theta_i=1)$ to the following quantity:
$$q(\theta_i=1) := \frac{q(\theta_i=1)}{q(\theta_i=1)+q(\theta_i=0)}.$$

Using the expression $q(\theta_i)$ as shorthand for $q(\theta_i=1)$, this results in the following update rule:%
\begin{align*}
    q(\theta_i=1) &= \sigma \Big(\log p(\theta_i) - \log (1-p(\theta_i)) \\ 
    &\phantom{=} - p_{\text{all\_off}} \cdot \log \frac{\alpha}{1-\alpha} \Big).
\end{align*}

In practice, we update over batches of inputs/outputs rather than single datapoints, \ie

\begin{align*}
\mathbf{m_{i,j}} = \sum_{j' \neq j \in \phi(x_i)} \log (1-p(\theta_j'))+\log{\log{\frac{\alpha}{1-\alpha}}},
\end{align*}
\begin{align*}
q(\theta_j) = \sigma(\log p(\theta_j) &- \log (1-p(\theta_j))\\
&-\sum_{i<t} y_i \cdot \exp(\mathbf{m_{i, j}})).
\end{align*}

We update each $q(\theta_j)$ either for a fixed number of steps $s$, or until convergence, \ie when:
\begin{align*}
\left|\sum_{j\in |\fullthetamath|}q^{\delta+1}_j-q^{\delta}_j\right| < \epsilon,
\end{align*}
where $\epsilon$ is an error threshold.

\end{document}